\documentclass[conference]{IEEEtran}
\IEEEoverridecommandlockouts
\IEEEoverridecommandlockouts
\usepackage{cite}
\usepackage{amsmath,amssymb,amsfonts}
\usepackage{algorithmic}
\usepackage{graphicx}
\usepackage{textcomp}
\usepackage{xcolor}
\usepackage[a4paper, total={184mm,239mm}]{geometry}
\def\BibTeX{{\rm B\kern-.05em{\sc i\kern-.025em b}\kern-.08em
    T\kern-.1667em\lower.7ex\hbox{E}\kern-.125emX}}

\newcommand{\HS}{\mathcal{H}} 
\newcommand{\bind}{\otimes} 
\newcommand{\bundle}{\oplus} 
\newcommand{\permute}{\Pi} 

\newenvironment{small-par-skip}
{\parskip=-2pt}

\newcommand{\name}[0]{HDCC}

\usepackage[hidelinks]{hyperref}
\usepackage{booktabs}
\usepackage{pgf}
\usepackage{import}
\usepackage{makecell}
\usepackage{tikz,tkz-euclide,pdftexcmds,calc}
\usetikzlibrary{arrows,shapes.geometric}
\usetikzlibrary{positioning}
\usetikzlibrary{automata}
\usetikzlibrary{arrows,shapes,calc}
\usepackage{circuitikz}
\usepackage{algorithmic}
\usepackage[linesnumbered,ruled,noend]{algorithm2e}
\usepackage{bbm}
\usepackage{multirow}

\usepackage{listings}

\begin{document}

\title{HDCC: A Hyperdimensional Computing compiler for classification on embedded systems and high-performance computing}

\author{Pere Vergés, Mike Heddes, Igor Nunes, Tony Givargis and Alexandru Nicolau\\
\IEEEauthorblockA{\textit{Department of Computer Science, University of California, Irvine
} \\
\textit{Irvine, California, United States of America}\\
(pvergesb, mheddes, igord, givargis, nicolau)@uci.edu}
}

\maketitle

\begin{abstract}
Hyperdimensional Computing (HDC) is a bio-inspired computing framework that has gained increasing attention, especially as a more efficient approach to machine learning (ML). This work introduces the \name{} compiler, the first open-source compiler that translates high-level descriptions of HDC classification methods into optimized C code. The code generated by the proposed compiler has three main features for embedded systems and High-Performance Computing: (1) it is self-contained and has no library or platform dependencies; (2) it supports multithreading and single instruction multiple data (SIMD) instructions using C intrinsics; (3) it is optimized for maximum performance and minimal memory usage. \name{} is designed like a modern compiler, featuring an intuitive and descriptive input language, an intermediate representation (IR), and a retargetable backend. This makes \name{} a valuable tool for research and applications exploring HDC for classification tasks on embedded systems and High-Performance Computing. 
To substantiate these claims, we conducted experiments with HDCC on several of the most popular datasets in the HDC literature. The experiments were run on four different machines, including different hyperparameter configurations, and the results were compared to a popular prototyping library built on PyTorch. The results show a training and inference speedup of up to 132$\times$, averaging 25$\times$ across all datasets and machines. Regarding memory usage, using 10240-dimensional hypervectors, the average reduction was 5$\times$, reaching up to 14$\times$. When considering vectors of 64 dimensions, the average reduction was 85$\times$, with a maximum of 158$\times$ less memory utilization.
\end{abstract}


\section{Introduction}
Hyperdimensional Computing (HDC)~\cite{kanerva2009hyperdimensional}, also known as Vector Symbolic Architectures (VSA)~\cite{gayler2003vector}, is a brain-inspired computation framework. In the context of machine learning, HDC is a resource-efficient alternative to deep learning. HDC uses high-dimensional vectors to encode and manipulate symbolic information, reducing memory and computational requirements. This makes the computing paradigm well-suited for on-the-fly data analysis in resource-constrained environments such as embedded systems where energy, memory, and processing power are limited. With its efficiency and scalability, HDC has become a promising solution for embedded machine learning applications, offering a low-power and resource-efficient alternative to deep learning\cite{lowpower}\cite{embeddedhdc}.
In HDC, data is represented as high-dimensional vectors (hypervectors) and arithmetic is performed by combining and comparing these hypervectors using similarity metrics.

HDC offers advantages such as smaller model sizes and lower computational costs compared to other machine learning approaches. As a result, HDC has become a promising solution for various applications, including natural language processing~\cite{rahimi2016robust}, time-series analysis~\cite{EgxClassification}, voice and gesture recognition~\cite{rahimi2016hyperdimensional}~\cite{imani2017voicehd}, graph learning~\cite{nunes2022graphhd}, and dynamic hashing techniques~\cite{heddes2022hyperdimensional}. These applications have demonstrated promising results in terms of accuracy and efficiency.

\begin{figure*}[htp!]
 \centering
 \caption{\label{fig:compiler-workflow} Workflow of the \name{} compilation and application execution.}
\includegraphics[width=\textwidth]{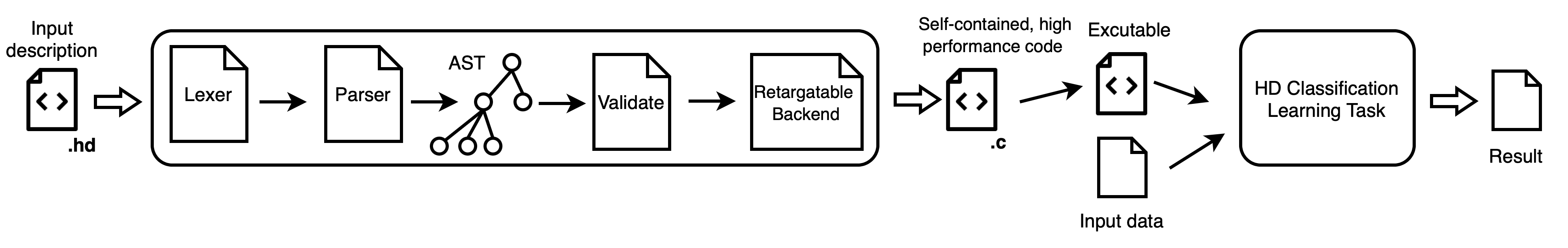}
\end{figure*}

Several high-level languages are used to build HDC classification models, including MATLAB~\cite{MATLAB:2010}, C++~\cite{ISO:2012:III} or Python~\cite{van1995python}. Within these, there are new high-level modeling frameworks, such as Torchhd~\cite{https://doi.org/10.48550/arxiv.2205.09208}, HDTorch~\cite{https://doi.org/10.48550/arxiv.2206.04746} and OpenHD~\cite{openhd}, among others. However, many languages and modeling frameworks require support from mathematical and ML libraries such as PyTorch. These tools are created for prototyping HDC models on Desktop environments or GPU platforms. Therefore, they may not suit the limitations of embedded systems due to their high demand for resources. In light of this, we propose a compiler to overcome these limitations, focused on providing a well-suited tool for execution on CPU-embedded environments. 
The compiler offers descriptive language (high level of abstraction). On its backend it applies resource optimizations that enable the execution of resource-constrained embedded systems with high performance and low peak memory usage. In Table~\ref{tab:average-time-gain}, we show the average and maximum time speedups of all four applications with different hyperspace dimensions where we reach an average and maximum speedup of 25$\times$ and 132$\times$ respectively with 10240 dimensional vectors. Moreover, in Table~\ref{tab:average-memory-gain}, we report the peak memory usage. We reach an average and maximum reduction of 5$\times$ and 14$\times$ on 10000 dimensions, respectively. 

\begin{table}[htp!]
  \centering
  \caption{\label{tab:average-time-gain} Average and maximum time speedup between all machines and datasets, compared to Torchhd.}
  \small
  \begin{tabular}{l|cccccc}
    \toprule
     \textbf{Dim} & \textbf{64} & \textbf{128} & \textbf{512} & \textbf{1024} & \textbf{4096} & \textbf{10240}\\
    \midrule
    Avg & 22.37x & 23.01x &   \textbf{28.00x} & 29.09x & 22.17x & 25.84x\\
    Max & 51.68x & 53.12x &  62.51x & 72.71x & 52.07x &  \textbf{132.61x}\\

    \bottomrule
\end{tabular}
\end{table}

\begin{table}[htp!]
  \centering
  \caption{\label{tab:average-memory-gain} Average and maximum memory reduction between all machines and datasets, compared to Torchhd.}
  \small
  \begin{tabular}{l|cccccc}
    \toprule
     \textbf{Dim} & \textbf{64} & \textbf{128} & \textbf{512} & \textbf{1024} & \textbf{4096} & \textbf{10240}\\
    \midrule
    Avg & \textbf{85.26x} & 76.87x & 49.20x & 30.17x & 10.29x & 5.83x\\
    Max & \textbf{158.67x} & 139.89x &  92.51x & 68.09x & 26.04x &  14.22x\\

    \bottomrule
\end{tabular}
\end{table}

The main contributions of this work are the following:
\begin{itemize}
    \item The \name{} compiler is a tool designed to take in a high-level, abstract description of a classification task, expressed in the form of a Hyperdimensional Computing (HDC) learning task. This input allows the user to specify the desired encoding for one-shot learning; the compiler generates efficient, self-contained HDC classification learning code optimized for multithreading, Single Instruction Multiple Data (SIMD) instructions using C intrinsics, and minimal memory usage. The resulting code is free of dependencies and suitable for use in resource-constrained environments such as embedded systems.
    \item \name{} is built as a modern compiler consisting of a tokenizer, grammar, validation, code generation, and retargetable backend.
    \item The \name{}\footnote{Project link: \url{https://anonymous.4open.science/r/hdcc-5F7C/}} compiler is open source and accessible to the community. All evaluations and results from this work, including code, data, and plots, are also available for complete reproducibility.
\end{itemize}

\section{Related Work}
Recently, the field of HDC has been the subject of exploration, and as a result, most researchers in this area have been creating their codes from scratch without relying on libraries. The most common programming languages used are Matlab and Python. To meet the need for faster prototyping, several tools have been developed to assist the HDC community.
The most popular tools are the following:

 \paragraph{Torchhd~\cite{https://doi.org/10.48550/arxiv.2205.09208}} a high-performance open-source library constructed on PyTorch. Its main objective is to facilitate the development process, making it easier for HDC experts to prototype models on desktop computers, and for non-experts to join the field. The framework is equipped to operate in both CPU and GPU environments.

 \paragraph{OpenHD~\cite{openhd}} a GPU-powered framework that uses high-performance computing (HPC) to streamline the implementation of HDC applications for classification and clustering on GPUs, providing high efficiency. The framework is written in Python and utilizes CUDA extensions, featuring integrated functions and decorators for the generation of HDC learning code.

 \paragraph{HDTorch~\cite{https://doi.org/10.48550/arxiv.2206.04746}} an open-source library for HDC, implemented in Python and incorporating CUDA extensions for hypervector operations. The aim of this framework is to simplify classical and online HDC learning, making it accessible to a wider audience through GPU support and CUDA compatibility. It is built upon the PyTorch library.

The aforementioned libraries are all constructed utilizing Python and demand various dependencies and system specifications. Their foremost purpose is to aid in the rapid prototyping of code, primarily on GPUs or desktop computers. However, \name{} differs from these libraries as it generates standalone C~\cite{kernighan2006c} code and does not require external libraries, thus having a minimal system and environment requirements. This attribute makes \name{} well-suited for easy development on embedded systems or systems with restricted resources. Furthermore, the descriptive language of \name{}, which is intuitive in nature, enables efficient exploration of the design space.

\section{Hyperdimensional Computing}
Hyperdimensional Computing has its roots in attempts to understand the workings of the brain and how information is represented in neurons~\cite{kanerva2014computing}. As a result of this research, several models of geometric spaces, known as hyperspaces, were developed. Consequently, there are various versions of the HDC framework, each with its own distinctive definition of the hyperspace and operations for encoding and processing symbolic information. Despite these variations, the common goal of all HDC methods is to provide a resource-efficient alternative to computing tasks, especially to traditional machine learning techniques, for embedded systems.

\subsection{The hyperspace}
There are several proposed variations of hyperdimensional computing models, including binary, real, and complex spaces, among others. Each of these spaces possesses its own distinct characteristics, operations, and metrics of similarity, allowing it to perform effectively within its defined space~\cite{schlegel2021comparison}. From now on, we assume the space is bipolar, i.e., $\HS = \{-1, +1\}^d$~\cite{gayler1998multiplicative}.
\subsection{The basis-hypervectors}
In HDC, hypervectors serve as the building blocks for encoding data within the hyperspace. Their design is inspired by the holographic principle, which leads to each dimension being identically and independently distributed. This results in hypervectors being robust and able to carry the same amount of information in each dimension. There are various types of basis-hypervectors, such as random-hypervectors that are randomly selected from the hyperspace, level hypervectors with a linear correlation, and circular hypervectors with a circular or cyclical correlation~\cite{nunes2022extension}. The high-dimensional nature of random-hypervectors leads them to be quasi-orthogonal~\cite{kanerva1988sparse}.
\subsection{Operations}
The arithmetic in HDC is performed using three highly parallelizable operations:
\begin{itemize}
    \item \textit{Binding} involves element-wise multiplication of two hypervectors, resulting in a hypervector that is dissimilar from the original ones.
    \begin{align*}
     \bind: \HS \times \HS \to \HS
    \end{align*}
    \item \textit{Bundling} involves the element-wise addition of two hypervectors, producing a hypervector that is similar to the original ones.
    \begin{align*}
     \oplus: \HS + \HS \to \HS
    \end{align*}
    \item \textit{Permuting} involves a cyclic shift of the hypervector. The shift is circular, meaning that the last element becomes the first if the shift is by one element.
    \begin{align*}
     \permute: \HS \to \HS
    \end{align*}
\end{itemize}

\subsection{Encoding}
The encoding of data in HDC is achieved by combining multiple hypervectors to form a more complex information representation. There are a variety of methods for encoding different forms of data, including text from characters~\cite{rahimi2016robust}, graphs from vertices and edges~\cite{nunes2022graphhd}, time series from samples~\cite{rahimi2016hyperdimensional}, and images from pixel values~\cite{manabat2019performance}. The subsequent section summarizes some of these established strategies.

\begin{itemize}
    \item A \textit{multiset} bundles together multiple hypervectors, creating a single hypervector that is similar to the inputs. 
    \begin{align*}
    Multiset(V_1, V_2, \dots, V_m) = V_1 \bundle V_2 \bundle \dots \bundle V_m
    \end{align*}
    \item The \textit{hash-table} encoding involves binding together two sets of hypervectors, one containing the keys and the other containing the corresponding values. These sets are then bundled to create a data structure that resembles a dictionary, where each value can be efficiently retrieved using its key.
    \begin{align*}
    HashTable(K_1, V_1, \dots, K_m, V_m) = \bigoplus_{i = 1}^{m} K_i \bind V_i   
    \end{align*}
    \item \textit{N-grams} are commonly utilized to represent sequences of letters, words, and sentences. This is achieved by combining $n$ consecutive hypervectors into a sliding window, followed by grouping the resulting hypervectors together.
    \begin{align*}
    Ngram(V_1, V_2, \dots, V_m) = \bigoplus_{i=1}^{m - n + 1} \bigotimes_{j = 0}^{n - 1}\permute^{n - j - 1}(V_{i + j})
    \end{align*}
    \item \textit{Graphs} are represented by binding vertices that share an edge and then bundling the resulting hypervectors as a multiset~\cite{nunes2022graphhd}.
\end{itemize}
These are some of the most common encoding patterns, but there are many others.

\subsection{Classification with HDC}
\paragraph{Training} involves creating a set of hypervectors containing the class prototype for each class. This set is placed in the associative memory during inference and is denoted as $M = \{C_1, C_2, \dots, C_k\}$. 
The prototype hypervector for each class is obtained by taking the element-wise majority vote of all the hypervectors in the input that belong to that class.
    \begin{align*}
    C_i = \bigoplus_{x : c(x) = i} \phi(x)
    \end{align*}
Each sample $x$ is encoded as the hypervector $\phi(x)$ and the element-wise majority vote of hypervectors is indicated by the $\bigoplus$ sign and is calculated using the bundling operation. The index $c(x)$ represents the class of the sample $x$.

\paragraph{Inference} this involves classifying given samples. First, the sample is encoded to a hypervector using the same encoding function as during training. The encoded hypervector is also known as a query-hypervector. This hypervector is then compared with the associative memory to determine which class it is most similar to.
\begin{align*}
     \hat{y} = argmax_i \, \delta (\phi(x), C_i)
\end{align*}
where $\delta(\cdot)$ denotes the similarity metric for the hyperspace, typically the cosine similarity.

\section{The \name{} Compiler}
This section describes the main components of the proposed compiler. These are comprised of the input language and its parser, consisting of a tokenizer, the grammar, and validation of the input, and the retargetable backend that generates the C code, which can generate either sequential or parallel code. The workflow of the \name{} compiler is illustrated in Figure~\ref{fig:compiler-workflow}.
\subsection{Language parser}
The compiler has as input a descriptive file of the desired configuration of the HDC learning classification task. These files contain a certain number of directives, such as the description of the embedding of the hypervectors used, the encoding of the learning task, and some other hyperparameters used for the execution, like the number of dimensions. An example of an input file is shown in Listing~\ref{code:voichd-decription}. Apart from the directives, one can also add comments in the code by adding ‘’//’’ after the directive, ending with ‘’;’’, which extends until the end of the line. In table~\ref{tab1}, we show the list of all available directives.

\begin{table}[htb]
\caption{The \name{} language specification}
\begin{center}
\begin{tabular}{|l|l|c|l|}
\hline 
\textbf{Directive} & \textbf{Parameters}& \textbf{Req}& \textbf{Description} \\
\hline
 .NAME & Str & Y & Output c file name.  \\ \hline
 .WEIGHT\_EMBED & Str Str Int & Y & Weight hypervector  \\ \hline
 .EMBEDDINGS & Str Str Int & N & Embeddings hypervectors  \\ \hline
.INPUT\_DIM & Int & Y & Input data dimension.  \\ \hline
 .ENCODING & Str & Y & Defines the hdc encoding. \\ \hline
 .CLASSES  & Int & Y & Number of classes. \\ \hline
 .TYPE  & Str & N & Type of code generation. \\ \hline
 .DIMENSIONS & Int & Y & Num of hdc dimensions. \\ \hline
 .TRAIN\_SIZE  & Int & Y & Num of training samples. \\ \hline
 .TEST\_SIZE  & Int & Y & Num of testing samples. \\ \hline
 .NUM\_THREADS  & Int & N & Number of threads used. \\ \hline
 .VECTOR\_SIZE  & Int & N & SIMD vector size \\ \hline
 .DEBUG  & Bool & N & Activates the debug logs. \\ \hline
\end{tabular} 
\label{tab1}
\end{center}
\end{table}

\label{code:voichd-decription}
\begin{lstlisting}[caption=Example of an HDCC description file for the VoiceHD~\cite{imani2017voicehd} classification task.]{text}

.NAME VOICEHD;
.WEIGHT_EMBED (VALUE LEVEL 100);
.EMBEDDING (ID RANDOM 617);
.INPUT_DIM 617;
.DEBUG TRUE;
.ENCODING MULTIBUNDLE(BATCHBIND(ID,VALUE));
.CLASSES 27;
.TYPE PARALLEL;
.DIMENSIONS 10240;
.TRAIN_SIZE 6238;
.TEST_SIZE 1559;
.VECTOR_SIZE 128;
.NUM_THREADS 4;
\end{lstlisting}

\subsection{HDC Learning Compiler}
\begin{figure}[h]
 \centering
 \caption{\label{fig:hdc-classification-workflow} \name{} classification learning task workflow.}
\includegraphics[width=\textwidth/2]{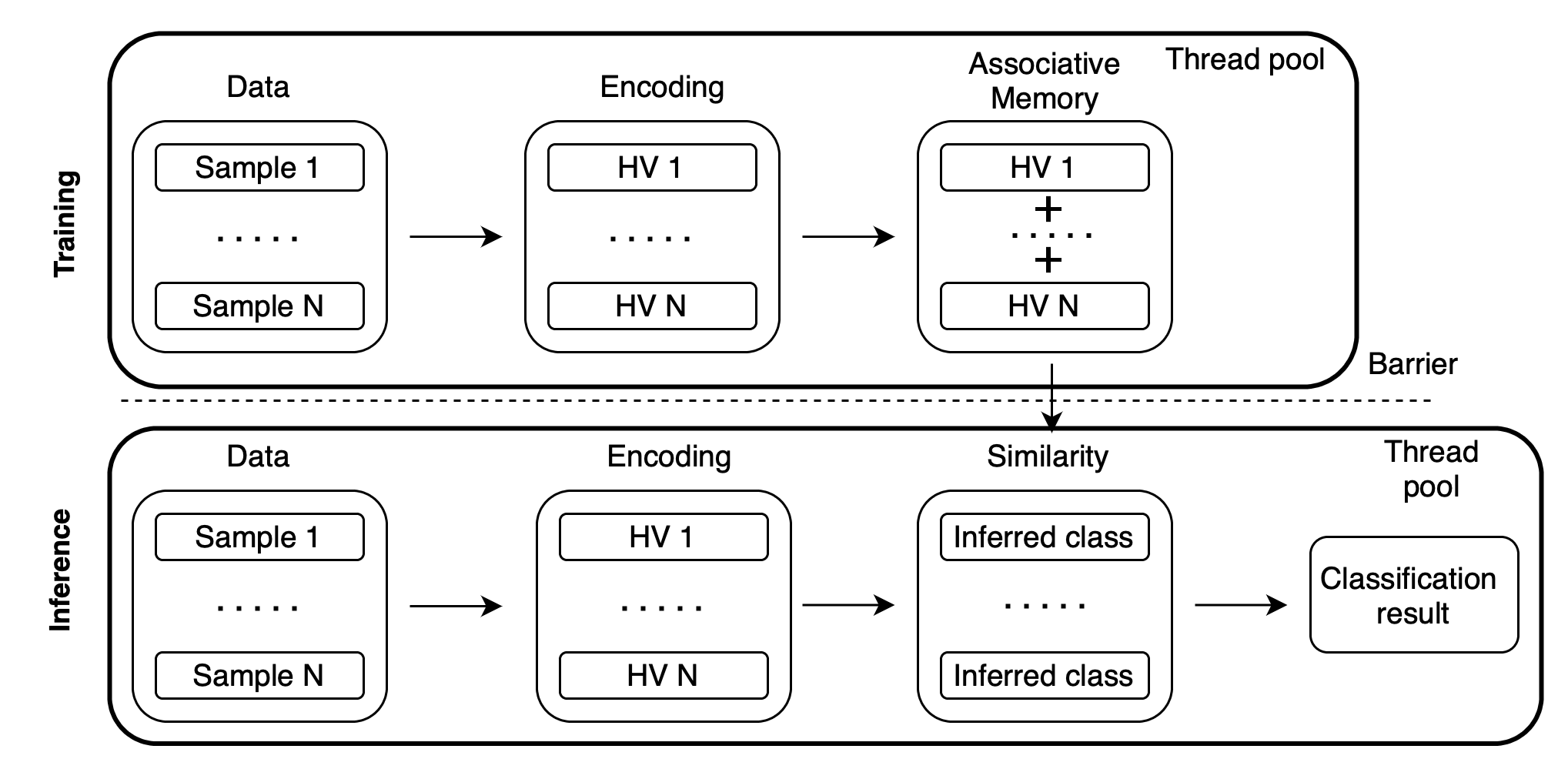}
\end{figure}

The \name{} compiler translates the program description described above into an HDC Learning model, which is composed of three different parts: encoding, training, and inference. 
The training performs one-shot learning by updating the associative memory, which contains one hypervector for each class in the classification task, with the bundling of the encodings of each data sample. Then, during inference, each input value will be encoded and compared based on the cosine similarity against the associative memory to assign its class. This process is shown in Figure~\ref{fig:hdc-classification-workflow}.

\subsubsection{Hyperspace representation}
This choice affects the type of operations and similarity metrics used. For the first version of the compiler we have chosen to implement the MAP (Multiply-Add-Permute) since it is one of the most commonly used by the community~\cite{kleyko2021survey1}. MAP uses element-wise addition and multiplication for bundling and binding, respectively.


\subsubsection{Embeddings}
The hypervectors, which in the instructions are referred to as embeddings, can have different initializations. The current implementation has two types of initialization: random, which is sampled uniformly from the hyperspace, the generated hypervectors will be quasi-orthogonal with high probability~\cite{kanerva1988sparse}; and level, which encodes linearly correlated information between two hypervectors that are uniformly sampled at random from the hyperspace. 

\subsubsection{Encoding}
The encoding is the most essential part of HDC one-shot learning. The \name{} translates the encoding from the description into a set of functions and encoding patterns, these functions are optimized to perform the minimum number of operations, these way optimizing the encoding given by the user. The encoding functions that have been implemented are the multiset, n-gram, and then the operations of bind and permute which can also be performed in batches to improve throughput.

\subsubsection{Training and inference}
The steps involved in the training phase for each sample are as follows: (1) read the data and normalize it, (2) assign hypervectors from the embeddings to feature values, (3) perform the specified encoding, (4) quantize the resulting hypervector, and (5) update the associative memory. This is a typical structure for HDC Learning training. For common patterns, such as combining bind and multiset or adjusting the weight embedding, \name{} has optimizations that consider both efficiency and memory usage, such as combining bind and multiset when used together, which use significantly less memory than performing them separately (on the order of input size times dimension).

The inference process follows a similar structure to the training process, which involves the following steps: (1) reading the data and normalizing it; (2) mapping each feature value to a hypervector from the embeddings; (3) performing the specified encoding; (4) quantizing the resulting hypervector; (5) probing the associative memory with the encoded sample by calculating the cosine similarity. An optimization has been made by normalizing the associative memory after training is complete; rather than calculating the cosine similarity for each test sample, only a dot product is needed, reducing the computation during inference.

A set of functions has been developed to carry out these steps, drawing inspiration from the API design of Torchhd \cite{https://doi.org/10.48550/arxiv.2205.09208}. These functions are stated in Table~\ref{training}.

\begin{table}[htb]
\caption{Set of functions implemented for HDC Learning.}
\begin{center}
\begin{tabular}{|l|c|}
\hline 
\textbf{Name} & \textbf{Description} \\
\hline
map\_range(input, indexes) & \makecell{Maps the input data into\\  the embedding range.} \\ \hline
forward(embedding, input) & \makecell{Assigns the correct embedding value \\to each value on the input data}   \\ \hline
encode & \makecell{Applies the encoding given by the user, \\ this function is in itself built by a set of \\ encoding functions mentioned previously.}   \\ \hline
hard\_quantize(encoding) & \makecell{Limits the output of the \\encoding to be binary.}   \\ \hline
update\_memory(encoding) & \makecell{Updates the associative with\\ the given encoding.}   \\ \hline
normalize() & \makecell{Normalizes the associative memory \\by applying the l2 norm.} \\\hline
linear(encoding) & \makecell{Applies a linear transformation to the \\ incoming data. $y = xA^T + b$} \\ \hline
argmax(hv) & \makecell{Returns the indices of the maximum \\value of the input hypervector.}  \\ \hline
\end{tabular} 
\label{training}
\end{center}
\end{table}

\subsection{SIMD: Vector intrinsics}
\name{} incorporates vector intrinsics in its computation processes, taking advantage of parallelism optimized by the compiler. PyTorch, a widely used framework for HDC, supports bit-level operations but is incapable of making addition operations on hypervectors. However, this limitation is not present in C vector intrinsics.
The user can adjust the value of this parameter as the optimal value may vary depending on the architecture. Some operations have been adapted to run with vector intrinsics, which require breaking down the hypervectors into the selected vector size for processing, as demonstrated in Listing~\ref{code:c-template-bind}. The utilization of vector intrinsics enables the efficient execution of HDC Learning tasks.


\label{code:intrinsics}
\begin{lstlisting}[
    basicstyle=\small, language=C, caption=Vector intrinsics definition]{c}
    
float v4si __attribute__((vector_size(128)));
\end{lstlisting}

\subsection{Multithreading}
\name{} has the capability to run as either a single-threaded or multithreaded system. The multithreaded option utilizes the standard API for C, the POSIX thread library, and is organized with a thread pool. The tasks in the thread pool consist of encoding a single instance, updating associative memory during training, and probing associative memory during inference.
This design was selected because it facilitates a single synchronization point prior to initiating inference. Moreover, relying on the POSIX thread to manage its resources as much as possible optimizes OS performance. The number of threads utilized during execution can be specified through the use of the .NUM\_THREADS directive.

\subsection{Memory optimization}
The generated C code by \name{} prioritizes minimizing memory usage. It only reads one data at a time, which will be as much as the number of threads used. The primary source of memory usage are the encoding operations. These, which are memory-intensive, can require memory equal to the input size multiplied by the size of the dimensions. However, \name{} optimizes the code by aggregating encoding and storing it in a constant space proportional to the dimensions, avoiding excessive memory usage.

\subsection{Retargetable backend}
The \name{} compiler, after successfully verifying the input description of the classification task, generates a series of C code functions that result in the application code. The compiler can be utilized to generate accelerated code for FPGA and/or GPU through the implementation of other backend outputs such as OpenCL or OpenGL.

\label{code:c-template-bind}
\begin{lstlisting}[
basicstyle=\small, language=C, caption=Vector intrinsics definition]{c}

f4si *batchbind(f4si *a, f4si *b){
    int i, j;
    f4si *enc = (f4si *)calloc(DIMENSIONS 
                * INPUT_DIM, sizeof(int));
    for(i = 0; i < INPUT_DIM; ++i){
        for(j = 0; j < NUM_BATCH; j++){
             enc[(NUM_BATCH * i) + j] = 
               a[(NUM_BATCH * i) + j] * 
               b[NUM_BATCH + j];
        }
    }
    return enc;
}
\end{lstlisting}

\subsection{Invoking the \name{}}
The \name{} compiler is freely accessible as a GIT repository and is developed in Python. It generates C code and has been successfully compiled on various operating systems such as macOS, Unix/Linux, and RISC-V. The only library utilized for producing the executables is PLY, which helps create the compiler's lexer and parser.

\begin{figure*}[h]
 \centering
 \caption{\label{fig:exp-results} Time, Accuracy, and Memory obtained by the execution of ISOLET, MNIST, EMG, and European Languages datasets by execution of the applications using the \name{} and Torchhd library (THD) in four different machines defined in Section~\ref{sec:machines} used for the experimental evaluation.}
\includegraphics[width=\textwidth]{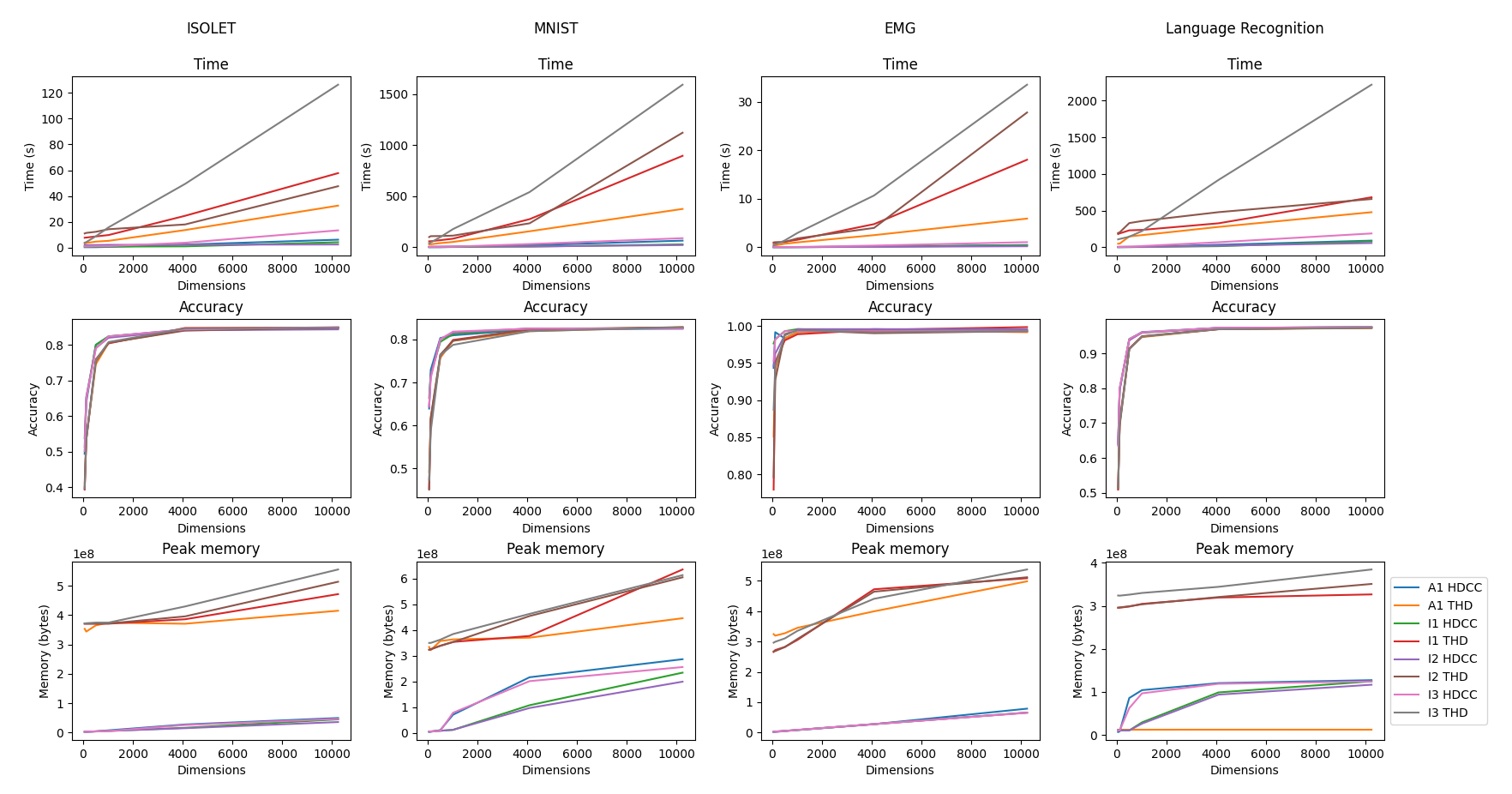}
\end{figure*}

The \name{} language enables the creation of an HDC Learning classification task, and the \name{} compiler will produce all the required files for compiling and executing the output files. A step-by-step guide on how to generate and execute a classification task using the ISOLET dataset (VoiceHD) will now be provided.
\begin{enumerate}
    \item Create the voicehd.hdcc \name{} description as it is shown in Listing \ref{code:voichd-decription}.
    \item Compile using:
        \begin{lstlisting}[
basicstyle=\small, language=bash]{bash}
python3 main.py voicehd.hdcc
        \end{lstlisting}
    \item The *.c and Makefile files are generated. Compile the *.c files using:
            \begin{lstlisting}{bash}
make
            \end{lstlisting}
    \item Run the application and pass in the data files as arguments. The data must be separated by commas for each value and each sample must be on a new line. All values must be either integers or floating-point numbers. To execute the compiled program use:
            \begin{lstlisting}[
basicstyle=\small, language=bash]{bash}

./voicehd train_data.txt train_labels.txt 
test_data.txt test_labels.txt 
            \end{lstlisting}

\end{enumerate}

\section{Experimental Results}
\subsection{Experimental Setup}
\subsubsection{Datasets and Applications}
For the experiments, we will utilize four datasets widely recognized by the Hyperdimensional Computing community.
\begin{itemize}
    \item ISOLET\cite{cole1990isolet}, a Speech Recognition dataset for classifying audio of the 26 English alphabet letters has been encoded using the method used in VoiceHD~\cite{imani2017voicehd}.
\label{code:isolet encoding}
            \begin{lstlisting}[
basicstyle=\small, language=bash, caption=ISOLET embedding and encoding.]{bash}

.WEIGHT_EMBED (VALUE LEVEL 100);
.EMBEDDING (ID RANDOM 617);
.ENCODING MULTIBUNDLE(
    BATCHBIND(ID,VALUE));
            \end{lstlisting}

    \item The EMG hand gesture recognition dataset features recordings of the hand position of five subjects, with five recorded positions to classify the data into. The encoding was obtained from~\cite{rahimi2016hyperdimensional}.
\label{code:emg encoding}
            \begin{lstlisting}[
basicstyle=\small, language=bash, caption=EMG embedding and encoding.]{bash}

.WEIGHT_EMBED (SIGNALS LEVEL 21);
.EMBEDDING (CHANNELS RANDOM 1024);
.ENCODING PERMUTE(MULTIBUNDLE(
    BATCHBIND(CHANNELS,SIGNALS)),1);
            \end{lstlisting}

  \item Language recognition~\cite{10.1145/2934583.2934624}, contains sentences from 21 European languages, sourced from Wortschatz Corpora which holds a vast number of sentences in the classified languages~\cite{10.1145/2934583.2934624}.
\label{code:langugae recognition}
            \begin{lstlisting}[
basicstyle=\small, language=bash, caption=Language recognition embedding and encoding.]{bash}

.WEIGHT_EMBED (SYMBOLS RANDOM 28);
.ENCODING NGRAM(SYMBOLS,3);
            \end{lstlisting}
    \item MNIST~\cite{deng2012mnist} this consists of images of handwritten digits, with each image representing a number from 0 to 9. The goal is to classify each image into one of the ten classes. 
\label{code:mnist encoding}
            \begin{lstlisting}[
basicstyle=\small, language=bash, caption=MNIST embedding and encoding.]{bash}

.WEIGHT_EMBED (POSITION LEVEL 1000);
.EMBEDDING (VALUE RANDOM 784);
.ENCODING MULTIBUNDLE(
    BATCHBIND(VALUE,POSITION));
            \end{lstlisting}
\end{itemize}

\subsubsection{Machines}
\label{sec:machines}
The experiments were conducted on four machines, with one using MacOS and the rest using UNIX OS. The machine specifications are as follows:
\begin{enumerate}
    \item Apple M2 chip (A1): 8-core CPU, eight threads per core, 16GB unified memory, 256GB SSD. Software MacOS Monterey. Python 3.9.13. Torchhd 4.0.0. Torch 1.13.0. Apple clang version 14.0.0.
    \item Intel(R) Xeon(R) CPU X5680 (I1): at 3.33GHz, with 24 cores, 2 threads per core. Software Ubuntu 22.04.1 LTS. Python 3.6.13. Torchhd 4.0.0. Torch 1.10.1. gcc (Ubuntu 11.3.0).
    \item Intel(R) Xeon(R) (I2) Silver 4114 CPU at 2.20GHz, with 40 cores, 2 threads per core. Software Ubuntu 22.04.1 LTS. Python 3.6.13. Torchhd 4.0.0. Torch 1.10.1. gcc (Ubuntu 11.3.0).
    \item Intel(R) Core(TM) (I3) i5-7200U CPU at 2.50GHz, with 4 cores, 2 threads per core. Software Ubuntu 20.04. Python 3.8.10. Torchhd 4.0.0. Torch 1.13.1. gcc (Ubuntu 9.4.0). Clang 15.0.7.
\end{enumerate}

\subsubsection{Evaluation tools}
In this study, the metrics evaluated were time and memory peak usage, as well as the accuracy of the applications as a validation of the code's correctness. The time execution was measured by considering only the train and test loops but excluding the data loading process for Torchhd. For our code generated with \name{}, the data loading is included in the time execution as it occurs during both the train and test loops. The time metric was obtained using the standard c time library for our code, and the python standard time library for Torchhd executions.

To measure the peak memory usage for the \name{} executions, we utilized the GNU "usr\textbackslash bin\textbackslash time -l" command. For Torchhd, we wanted to obtain the actual peak memory usage, as opposed to the allocated memory, as python may allocate more memory than what is actually used. Hence, we used the memory-profiler tool~\cite{pyprofiler} to determine the exact peak memory usage. This was done by running the "mprof run" command during the application execution.

\begin{table}[h]
  \centering
  \caption{\label{tab:isolet-results} ISOLET experimental results using 10240 dimensions and the Intel2 machine.}
  \small
  \begin{tabular}{l|cc}
    \toprule
     \textbf{Metrics} & \textbf{Torchhd} & \textbf{HDCC}\\
    \midrule
    Time & 47.56s & 2.29s \textbf{(20.70x)}\\
    Memory & 513 MiB  & 36 MiB \textbf{(14.22x)}\\
    Accuracy & 85.2\% & 84.8\%\\
    \bottomrule
\end{tabular}
\end{table}

\begin{table}[h]
  \centering
  \caption{\label{tab:mnist-results} MNIST experimental results using 10240 dimensions and the Intel2 machine.}
  \small
  \begin{tabular}{l|cc}
    \toprule
     \textbf{Metrics} & \textbf{Torchhd} & \textbf{HDCC}\\
    \midrule
    Time & 49.96s & 1122s \textbf{(22.4x)}\\
    Memory & 605 MiB & 199 MiB \textbf{(3.03x)}\\
    Accuracy & 82.7\% & 82.8\%\\
    \bottomrule
\end{tabular}
\end{table}

\begin{table}[h]
  \centering
  \caption{\label{tab:emg-results} EMG experimental results using 10240 dimensions and the Intel2 machine.}
  \small
  \begin{tabular}{l|cc}
    \toprule
     \textbf{Metrics} & \textbf{Torchhd} & \textbf{HDCC}\\
    \midrule
    Time & 27.79s & 0.2s \textbf{(132.61x)}\\
    Memory & 511 MiB & 65 MiB \textbf{(7.77x)}\\
    Accuracy & 100\% & 99.3\%\\
    \bottomrule
\end{tabular}
\end{table}

\begin{table}[h]
  \centering
  \caption{\label{tab:lang-results} European Languages experimental results 10240 dimensions and using the Intel2 machine.}
  \small
  \begin{tabular}{l|cc}
    \toprule
     \textbf{Metrics} & \textbf{Torchhd} & \textbf{HDCC}\\
    \midrule
    Time & 657.69s & 58.06s \textbf{(11.32x)}\\
    Memory & 350 MiB & 116 MiB \textbf{(3.00x)}\\
    Accuracy & 97.2\% & 97.4\%\\
    \bottomrule
\end{tabular}
\end{table}

\subsubsection{Library comparison}
For performance evaluation, we conducted experiments using our \name{} and Torchhd libraries. As far as we know, there are no other HDC libraries that allow execution on the CPU, despite some claiming to, when using them, it was discovered that cuda was necessary to execute even when selecting the CPU option. Our \name{} implementations strictly follow the embeddings and encoding used in the Torchhd library.

The experiments were conducted with dimensions of 64, 128, 512, 1024, 4096, and 10240. For each application and dimension, the experiments were repeated five times, and the results were averaged.

In the experiments with \name{}, machines A1 and I3 used 4 threads, while machines I1 and I2 used 20 threads. This number was determined through experimentation and was found to provide the best results for each machine.

\subsubsection{Results}
The results of all experiments are displayed in Figure \ref{fig:exp-results}. The accuracy is almost identical between Torchhd and \name{}. The execution time with Torchhd increases more quickly than with \name{} when using more dimensions. Table \ref{tab:emg-results} shows a 132x speed improvement when using \name{} to run the EMG application with 10,240 dimensions on machine I2. The results also indicate a reduction in peak memory usage, as demonstrated in Table \ref{tab:isolet-results}, where memory usage was reduced by up to 14x when using 10,240 dimensions.

\section{Future Work}
The \name{} is a fully functional compiler that allows the execution of HDC Learning algorithms and easy design space exploration for embedded systems. The plan for enhancing the tool includes adding various VSA models (BSC~\cite{bsc} and FHRR~\cite{hrr}), advanced embedding functions (e.g. graph embeddings), and GPU/FPGA backend implementations to augment performance and functionality. 

\section{Conclusions}
This paper introduces the \name{} compiler, which is an open-source solution for translating Hyperdimensional Computing (HDC) Learning applications into C code for embedded systems. Designed like a modern compiler, it features an intuitive and descriptive input language, an Intermediate Representation (IR), and a flexible backend. The resulting C code is self-contained and requires minimal dependencies, making it suitable for systems with limited resources. We have shown that this compiler can be used in different environments. A comparison against Torchhd, the state-of-the-art HD library, implemented in Pytorch, has been done.
\name{} has outperformed in time and peak memory usage on all four machines and on the datasets tested, SOLET, EMG gestures, European Languages, and MNIST. The results on 10,000 dimensions achieved an average improvement in speed across all datasets and machines of 25$\times$, reaching a maximum of 132$\times$ faster than Torchhd during training and inference. On 10000 dimensions, there is an average memory reduction of 5$\times$ and a maximum reduction of 14$\times$.

\bibliographystyle{IEEEtran}
\bibliography{IEEEabrv,IEEEexample}

\end{document}